\documentclass[10pt,twocolumn,letterpaper]{article}

\usepackage{ijcb}
\usepackage{times}
\usepackage{epsfig}
\usepackage{graphicx}
\usepackage{amsmath}
\usepackage{amssymb}

\usepackage{authblk}

\ijcbfinalcopy 

\def\ijcbPaperID{138} 

\ifijcbfinal\fi
\begin{document}

\title{Occlusion-Adaptive Deep Network for Robust Facial Expression Recognition}

\author{Hui Ding
}
\author{Peng Zhou}
\author{Rama Chellappa}
\affil{University of Maryland, College Park}

\maketitle
\thispagestyle{empty}

\begin{abstract}
Recognizing the expressions of partially occluded faces is a challenging computer vision problem. Previous expression recognition methods, either overlooked this issue or resolved it using extreme assumptions. Motivated by the fact that the human visual system is adept at ignoring the occlusion and focus on non-occluded facial areas, we propose a landmark-guided attention branch to find and discard corrupted features from occluded regions so that they are not used for recognition. An attention map is first generated to indicate if a specific facial part is occluded and guide our model to attend to non-occluded regions. 
To further improve robustness, we propose a facial region branch to partition the feature maps into non-overlapping facial blocks and task each block to predict the expression independently. 
This results in more diverse and discriminative features, 
enabling the expression recognition system to recover even though the face is partially occluded. 
Depending on the synergistic effects of the two branches, our occlusion-adaptive deep network significantly outperforms state-of-the-art methods on two challenging in-the-wild benchmark datasets and three real-world occluded expression datasets.

\end{abstract}

\section{Introduction}
Facial expressions play an important role in social communication in our daily life.
In recent years, automatically recognizing expression has received increasing attention due to its wide applications, including driver safety, health care, video conferencing, virtual reality, and cognitive science \emph{etc}. 


Existing methods that address expression recognition can be divided into two categories. 
One category utilized synthesis techniques to facilitate discriminative feature learning~\cite{yang2018facial,zhang2018joint,bozorgtabar2019using,cai2019identity}; While the other tried to boost the performance by designing new loss functions or network architectures~\cite{li2017reliable,zeng2018facial,cai2018island,acharya2018covariance}.
In the first category, de-expression residue learning proposed in~\cite{yang2018facial} leveraged the neutral face images to distill the expression information from the corresponding expressive images. Zhang \etal~\cite{zhang2018joint} explored an adversarial autoencoder to generate facial images with different expressions under arbitrary poses to enlarge the training set. However, those works mainly focus on lab-collected datasets captured in controlled environments, such as CK+~\cite{lucey2010extended}, MMI~\cite{valstar2010induced} and OULU-CASIA~\cite{zhao2011facial}. Although high accuracy results have been obtained on these datasets, they perform poorly when recognizing facial expressions in-the-wild.
In the second category, Li \etal~\cite{li2017reliable} proposed a locality preserving loss to enhance deep features
by preserving the locality closeness measure while maximizing the inter-class scatters. To address the annotation inconsistencies among different facial expression datasets, Zeng \etal~\cite{zeng2018facial} introduced a probability transition layer to recover the latent truths from noisy labels. Although expression datasets under natural and uncontrollable conditions are explored, facial expression recognition under partial occlusions is still a challenging problem that has been relatively unexplored. In real-life images or videos, facial occlusions can often be observed, \emph{e.g.}, facial accessories including sunglasses, scarves, and masks or other random objects like hands, hairs and cups. 

Recently, some related works have been proposed to solve this challenge. 
Patch-gated Convolutional Neural Network~\cite{li2018patch}
decomposed a face into different patches and explicitly predicted the occlusion likelihood of the corresponding patch using a patch-gated unit. 
Wang \etal~\cite{wang2020region} proposed a self-attention scheme to learn the importance weights for multiple facial regions.
However, 
the unobstructed scores are learned without any ground truth on the occlusion information and may be biased. 
In this work, we present an \textbf{Occlusion-Adaptive Deep Network} (OADN) to overcome the occlusion problem for robust facial expression recognition in-the-wild. It consists of two branches: a landmark-guided attention branch and a facial region branch. 

The landmark-guided attention branch is proposed to discard feature elements that have been corrupted by occlusions. The interest points covering the most distinctive facial areas for facial expression recognition are computed based on the domain knowledge. Then the meta information of these points is utilized to generate the attention maps.
The global features are modulated by the attention maps to guide the model to focus on the non-occluded facial regions and filter out the occluded regions. 

To further enhance robustness and learn complementary context information, we introduce a facial region branch to train multiple region-based expression classifiers. This is achieved by first partitioning the global feature maps into non-overlapping facial blocks. Then each block is trained by backpropgating the recognition loss independently.
Thus even when the face is partially occluded, the classifiers from other non-occluded regions are still able to function properly. Furthermore, since the expression datasets are usually small, having multiple region-based classifiers adds more supervision and acts as a regularizer to alleviate the overfitting issue.


The main contributions in this work are summarized as follows: 
\begin{itemize}
\item We propose OADN, an effective method to deal with the occlusion problem for facial expression recognition in-the-wild. 
\item We introduce a landmark-guided attention branch to guide the network to attend to  non-occluded regions for representation learning.
\item We design a facial region branch to learn region-based classifiers for complementary context features and further increasing the robustness.
\item Experimental results on five challenging benchmark datasets
show that our proposed OADN obtains significantly better performance
than existing methods.
\end{itemize}

\section{Related Work}
\subsection{Deep Facial Expression Recognition}
Deep learning methods~\cite{zhong2012learning,liu2013aware,liu2014feature,liu2014facial,liu2014learning,yang2018facial,cai2019identity,zhang2018joint,li2017reliable,cai2018island,zeng2018facial,luo2018local,acharya2018covariance} for facial expression recognition have achieved great success in the past few years.
Based on the assumption that a facial expression is the combination of a neutral face image and the expressive component, Yang \etal~\cite{yang2018facial} proposed a de-expression residue learning to learn the residual expressive component in a generative model. 
To reduce the
inter-subject variations, Cai \etal~\cite{cai2019identity} introduced an identity-free generative adversarial network~\cite{goodfellow2014generative} to generate an average identity face image while keep the expression unchanged. Considering the pose variation, Zhang \etal~\cite{zhang2018joint} leveraged an adversarial autoencoder to augment the training set with face images under different expression and poses.   
However, these methods mainly focused on datasets captured in controlled environments, where the facial images are near frontal. Thus the models generalize poorly when recognizing human expressions under natural and uncontrollable variations.

Another line of work focused on designing advanced network architectures~\cite{acharya2018covariance} or loss functions~\cite{li2017reliable,cai2018island,zeng2018facial,luo2018local}. Li \etal~\cite{li2017reliable} proposed a deep locality-preserving Convolutional Neural Network, which preserved the local proximity by minimizing the distance to K-nearest neighbors within the same class. Building on this, Cai \etal~\cite{cai2018island} further introduced an island loss to simultaneously reduce intra-class variations and augment inter-class differences. Zeng \etal~\cite{zeng2018facial} studied the annotation error and bias problem among different facial expression datasets. 
Each image is predicted with multiple pseudo labels and a model is learned to fit the latent truth from these inconsistent labels. 
Acharya \etal~\cite{acharya2018covariance} explored a covariance pooling layer to better capture the distortions in
regional facial features and temporal evolution of per-frame features. Although the aforementioned approaches achieve good performance on data in the wild, facial expression recognition is still challenging due to the existence of partially occluded faces.

\subsection{Occlusive Facial Expression Recognition}
Recently, there are some works starting to investigate the occlusion issue. Li \etal~\cite{li2018occlusion} proposed a gate unit to enable the model to shift attention from the occluded patches to other visible facial regions. The gate unit estimates how informative a face patch is through an attention net, then the features are modulated by the learned weights. Similarly, region attention network~\cite{wang2020region} cropped multiple face regions and utilized a self-attention based model to learn an important weight for each region. However, the self-attention based methods lack additional supervision to ensure the functionality. Thus, the network may not be able to locate these non-occluded facial regions accurately under large occlusions and poses.

\section{Occlusion Adaptive Deep Network}
\begin{figure*}
  \includegraphics[width=\textwidth,height=0.45\textwidth]{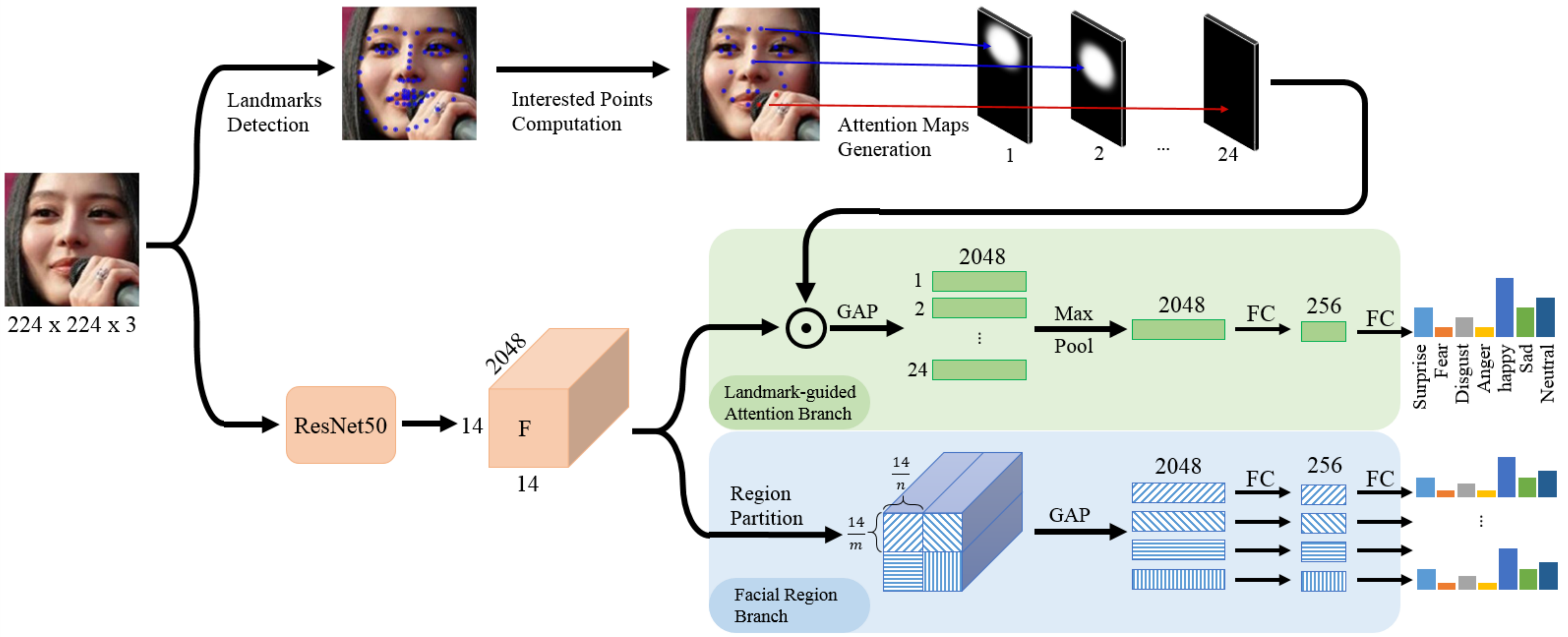}
  \caption{Pipeline of the Occlusion Adaptive Deep Network. It consists of two branches: a Landmark-guided Attention Branch and a Facial Region Branch. The ResNet50 backbone is shared between the two branches to extract the global features.
For the Landmark-guided Attention Branch, the facial landmarks are first detected. Then the interest points are computed to cover the most informative facial areas. The confidence scores of these points are further utilized to generate the attention maps, guiding the model to attend to the visible facial components. For the Facial Region Branch, the feature maps are divided into non-overlapping facial blocks and each block is trained to be a discriminative expression classifier on its own.}
  \label{fig:oadn}
\end{figure*}

To this end, we propose OADN for robust facial expression recognition in-the-wild. 
To be specific, we use ResNet50~\cite{he2016deep} without the average pooling layer and the fully connected layer as the backbone to extract global feature maps from given images. 
We set the stride of conv4\_1 to be 1, so a larger feature map is obtained. For an input image with height $H$ and width $W$, the resolution of the output feature $F$ will be $H/16\times W/16$ instead of $H/32\times W/32$. This is beneficial to identify the occlusion information and focus on the visible facial regions.  

As illustrated in Figure~\ref{fig:oadn}, OADN mainly consists of two branches: one is the landmark-guided attention branch, which utilizes a landmark detector to locate the landmarks and to guide the network to attend to the non-occluded facial areas. The other one is the facial region branch which divides the global feature maps into blocks and utilizes region-based classifiers to increase robustness.
 We describe each branch and the structural relationship among the two branches below.

\begin{figure}[!ht]
  \centering
  \includegraphics[width=0.4\textwidth,height=0.16\textwidth]{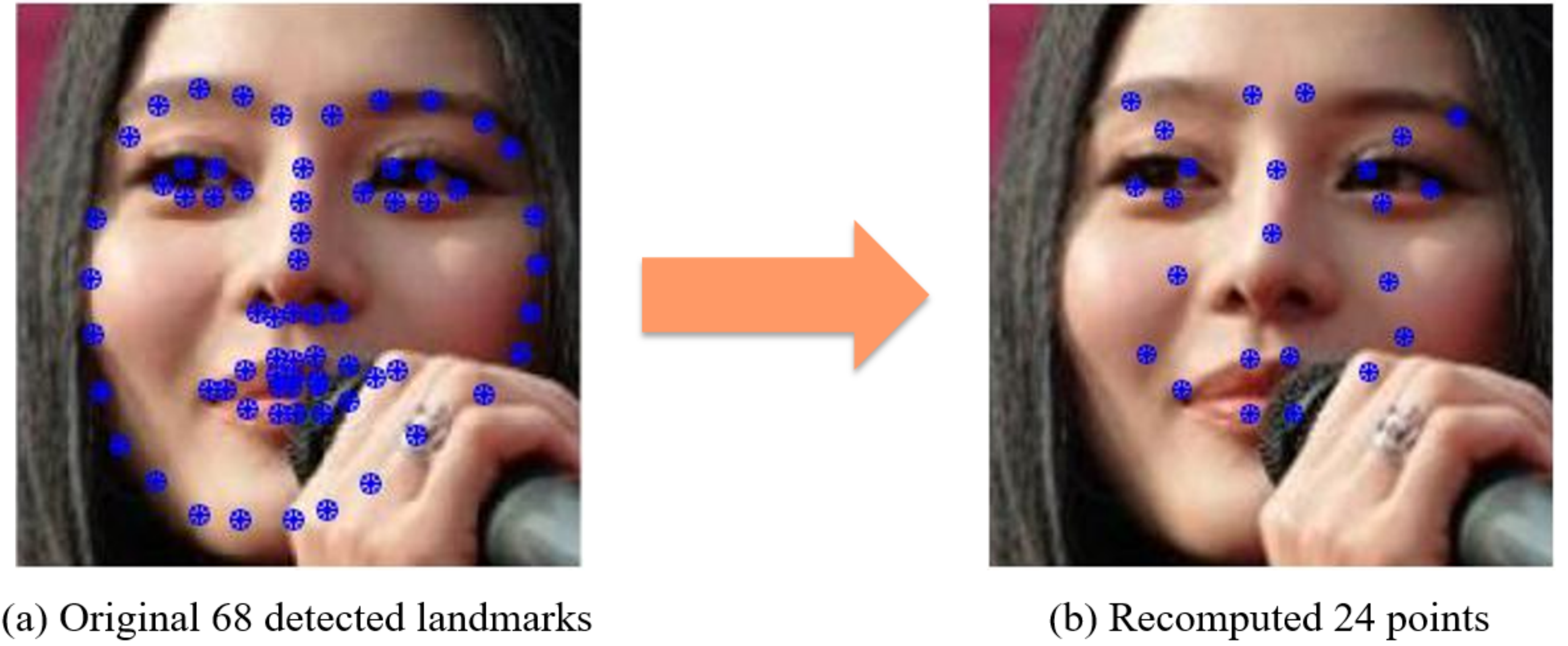}
  \caption{We select 16 points from the original 68 landmarks (a) to cover the regions around eyes, eyebrows, nose and mouth. We further recompute 8 points (b) to cover facial cheeks and the areas between eyes and eyebrows.}
  \label{fig:24pts}
\end{figure}

\subsection{Landmark-guided Attention Branch}
OADN employs a facial landmark detector~\cite{dong2018style} to obtain landmarks from face images. The landmark detector is pre-trained on the 300W dataset~\cite{sagonas2013300}. Given an input image, OADN utilizes the detector to extract $N=68$ landmarks. For each landmark, the detector predicts its coordinates and confidence score. Then based on the detected 68 points, we select or recompute $M=24$ interest points that cover the distinctive regions of face, including the eyes, nose, mouth and cheeks. Figure~\ref{fig:24pts} illustrates the computation results. For those recomputed points (mainly around eyes and cheeks), we set their confidence scores to be the minimum confidence score of landmark points that used to compute them. To remove the occluded facial regions, we set a threshold $T$ to filter out the landmarks that have confidence scores smaller than $T$. Specifically, the interest points are obtained by: 
\begin{equation}
p_i=\left\{
\begin{array}{lcl}
(x_i, y_i)      &      & {if~~s_i^{conf} \geq T}\\
0     &      & {else}
\end{array} \right. 
\end{equation}

where $p_i$ denotes the $i$th interest point, and $x_i$, $y_i$ denote the coordinates of the $i$th point. $s_i$ is the confidence score ranged from 0 to 1 and $T$ is the threshold.

We then generate the attention heatmaps consisting of a 2D Gaussian distribution, where the centers are the ground truth locations of the visible landmarks.
For those occluded landmarks, the corresponding attention maps are set to zero. We further downsample the attention maps by linear interpolation to match the size of the output feature maps. As shown in Figure~\ref{fig:oadn}, the attention map $A_i$ modulates the global feature maps $F$ to obtain the re-weighted features $F_i^A$. To achieve this, the feature map $F$ from the backbone is multiplied by each attention map $A_i,~i=1,...,M$ element-wisely, resulting  $M$ landmark-guided feature maps $F_i^A$: 
\begin{equation}
    F_i^A = F \odot A_i, i=1,...,M
\end{equation}
where $A_i$ is the $i$th heatmap, and $\odot$ is element-wise product.
Since the attention map indicates the visibility of each facial component, the landmark-guided feature map $F_i^A$ can attend to the non-occluded facial parts and remove the information from the occluded regions. Thus, the feature from the visible region is emphasized and occluded part is canceled. 

Then global average pooling is applied to each landmark-guided feature map $F_i^A$ to obtain a 2048-$D$ feature $f_i^A, i=1,...,M$, corresponding to the facial component containing the specific interest point. Finally, the component-wise feature $f_i^A$ is max-pooled to fuse features from the non-occluded facial areas. A fully-connected layer is further used to reduce the dimension from 2048 to 256, and the output is fed into a softmax layer to predict the expression category of each input face image. We utilize the cross-entropy loss to train the landmark-guided attention branch, which is expressed as follows:
\begin{equation}
    L_{LAB} = -\sum_{i=1}^Cy_i\log\hat{y_i}
\end{equation}
where $\hat{y_i}$ is the prediction, $y_i$ is the ground truth and $C$ is the number of expression classes.

\subsection{Facial Region Branch}
When the face is severely occluded, the landmark detection results may not be accurate. Since relying on the landmark-guided attention branch solely is not enough,
OADN utilizes a Facial Region Branch (FRB) to learn useful context information and further increase the robustness.

Given the global feature maps $F\in h\times w\times c$, where $h, w, c$ are the height, width and channel dimensions, we first divide them into small $m\times n$ non-overlapping blocks.
Each facial region feature $F_i^R\in m\times n \times c, i=1,...,K$, where $K = \lceil{\frac{h}{m}}\rceil \cdot \lceil{\frac{w}{n}}\rceil$ is then fed into a global average pooling layer to obtain a region-level feature $f_i^R$. Afterwards, a fully-connected layer is employed to reduce the dimension of $f_i^R$ from 2048 to 256. 
Finally, a softmax layer is applied to each region to obtain a set of predictions $y_i^R$, where $i=1,...,K$.

To train the facial region branch, we minimize the cross-entropy loss over the $K$ regions independently. Formally, the loss is expressed as:
\begin{equation}
    L_{FRB} = -\sum_{i=1}^C\sum_{j=1}^Ky_i\log\hat{y}_{i,j}^R
\end{equation}
where $K$ is the number of facial regions, $\hat{y}_{i,j}^R$ is the prediction of the $j$th region, and $y_i$ is the ground truth expression category. 

To make an accurate prediction based on facial region only, OADN is required to learn more discriminative and diverse features at a finer-level. As a result, the partial occlusion will have a less effect on the network compared to a standard model. Moreover, the size of the expression recognition dataset is usually not very large. Training multiple region-based classifiers adds more supervision and reduces overfitting. 

\subsection{Relationship between the Two Branches}
OADN is specifically designed to handle the occlusion problem for in-the-wild facial expression recognition.
The landmark-guided attention branch explicitly guides the model to focus on non-occluded facial areas, learning a clean global feature. While the facial region branch promotes part-level features and enables the model to work robustly when the face is largely occluded. 
Combining the benefits from each branch, we train OADN using the following loss:
\begin{equation}
L = \lambda L_{LAB} + (1-\lambda) L_{FRB}
\end{equation}
where $\lambda$ is the loss combination weight.
$L_{LAB}$ and $L_{FRB}$ are defined in Equation (3) and (4).

\section{Experiments}
\subsection{Datasets}
We validate the effectiveness of our method on two largest in-the-wild expression datasets: RAF-DB~\cite{li2017reliable} and AffectNet~\cite{mollahosseini2017affectnet}. The in-the-wild datasets contain facial expression in real world with various poses, illuminations, intensities, and other uncontrolled conditions. We also evaluate our method on three recently proposed real-world occlusion datasets: Occlusion-AffectNet\cite{wang2020region}, Occlusion-FERPlus~\cite{wang2020region} and FED-RO~\cite{li2018occlusion}. The occlusions are diverse in color, shape, position and occlusion ratio.

\begin{figure}[!ht]
  \centering
  \includegraphics[width=0.48\textwidth,height=0.1\textwidth]{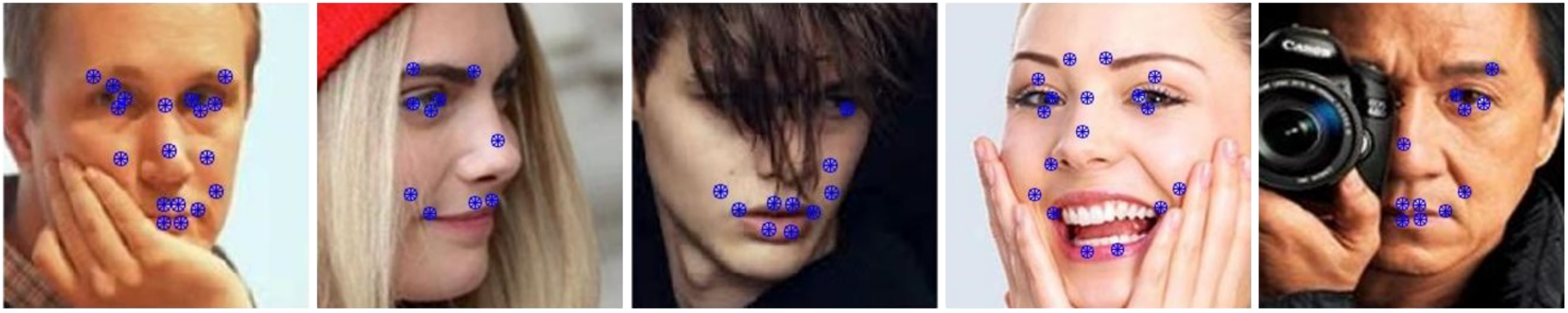}
  \caption{The interest points with confidence scores greater than the threshold $T$ are shown in red points. We can see that occluded facial areas are removed.
   }
  \label{fig:conf_score}
\end{figure}

\subsection{Implementation Details}
\textbf{Preprocessing.} The standard MTCNN~\cite{zhang2016joint} is used to detect five face landmarks for all the images. After performing a similarity transformation, we obtain the aligned
face images and resize them to be 224 x 224 pixels. To detect landmarks from occluded images, we use SAN~\cite{dong2018style} pre-trained on the 300W dataset~\cite{sagonas2013300} to get 68 face landmarks. We also try another landmark detector~\cite{bulat2017far} and similar results are obtained. Then we select 18 points covering eyebrows, eyes, nose and mouth, and recompute eight points from facial cheeks. The confidence scores of these recomputed points are the minimum scores of the points used to compute them. In all experiments, we set the threshold $T$ of the confidence score to be 0.6, thus landmarks with confidence scores smaller than 0.6 are removed. Figure~\ref{fig:conf_score} shows the computed interest points after thresholding. From it we can see that the occluded facial regions are discarded. Finally, we generate attention maps consisting of a Gaussian with the centers as the coordinates of the visible points. For those occluded points, the attention maps are all zeros. We resize the attention maps to be $14\times 14$ to match the size of the global feature maps $F$.

\textbf{Training and Testing.} We employ the ResNet50 as our backbone, removing the average pooling layer and the fully connected layer. We modify the stride of conv4\_1 from 2 to 1, so a larger feature map with size $14\times 14$ is obtained. We initialize the model with the weights pre-trained on ImageNet~\cite{deng2009imagenet}. The mini-batch size is set to be 128, the momentum is 0.9, and the weight decay is 0.0005. The learning rate starts at 0.1, and decreased by 10 after 20 epochs. We train the model for a total of 60 epochs. Stochastic Gradient Descent (SGD) is adopted as the optimization algorithm. During training, only random flipping is used for data augmentation.
For testing, a single image is used and the predication scores from the landmark-guided attention branch and the facial region branch are averaged to get the final prediction score. The settings are same for all the experiments. For evaluation, the total accuracy metric is adopted. Considering the imbalance of the expression classes, a confusion matrix is also employed to show the average class accuracy.
The deep learning software Pytorch~\cite{paszke2019pytorch} is used to conduct the experiments. Upon publication, the codes and trained
expression models will be made publicly available.

\begin{table}
\caption{Test set accuracy on RAF dataset}
\label{label:raf_acc}
\begin{center}
\begin{tabular}{c|c}
\hline
Method & Average Accuracy \\
\hline
\hline
RAN~\cite{wang2020region} & 86.90\% \\
\textbf{OADN(ours)}  & \textbf{89.83\%}\\
\hline
ResiDen~\cite{jyoti2019expression} & 76.54\% \\
ResNet-PL~\cite{pan2019occluded} & 81.97\%  \\
PG-CNN~\cite{li2018patch} & 83.27\% \\
Center Loss~\cite{wen2016discriminative} & 83.68\%  \\
DLP-CNN~\cite{li2018reliable} & 84.13\%  \\
ALT~\cite{floreaannealed} & 84.50\% \\
gACNN~\cite{li2018occlusion} & 85.07\%   \\
\textbf{OADN(ours)}  & \textbf{87.16\%}\\
\hline
\end{tabular}
\end{center}
\end{table}
\subsection{Results Comparison}
\textbf{RAF}~\cite{li2017reliable} contains 30,000 in-the-wild
facial expression images, annotated with basic or compound
expressions by forty independent human labelers. In this experiment, only images with seven basic expressions are used, including 12,271 for training and 3,068 for testing.

Table~\ref{label:raf_acc} shows the results of our method and previous works. 
Our OADN achieves 87.16\% in terms of total accuracy on the test set, outperforming all the previous methods. Compared with the strongest competing method in the same setting gACNN~\cite{li2018occlusion}, OADN surpasses it by 2.1\%. This is because OADN explicitly utilizes the meta information of landmarks to depress the noisy information from the occluded regions and enhances the robustness with multiple region-based classifiers. 
To have a fair comparison with~\cite{wang2020region}, we also pre-trained our model on a large-scale face recognition dataset VGGFace2~\cite{cao2018vggface2}. OADN achieves a new state-of-the-art result with an accuracy of 89.83\% to the best of our knowledge, outperforming RAN by 2.93\%. 
This validates the superiority of the proposed method.

\begin{figure}[!ht]
  \centering
  \includegraphics[width=0.45\textwidth]{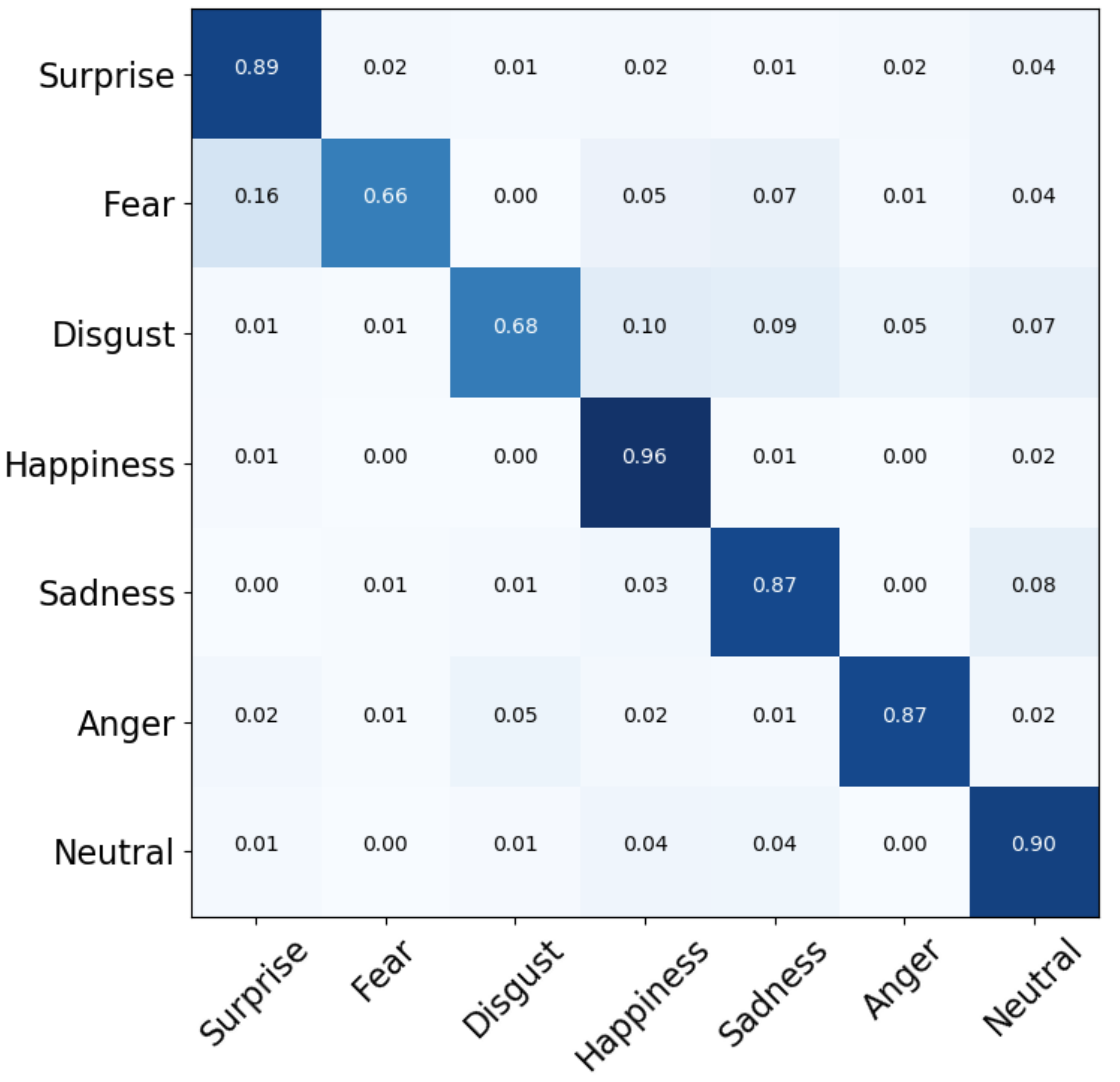}
  \caption{Confusion matrix for RAF-DB dataset. The darker the color, the higher the accuracy.}
  \label{fig:raf_cm}
\end{figure}


\begin{table}
\caption{Validation set accuracy on AffectNet dataset}
\label{tab:affect_acc}
\begin{center}
\begin{tabular}{c|c}
\hline
Method & Average Accuracy   \\
\hline
\hline
RAN~\cite{wang2020region} & 59.50\% \\
\textbf{OADN(ours)}  & \textbf{64.06\%} \\
\hline
VGG16~\cite{simonyan2014very} & 51.11\%  \\
GAN-Inpainting~\cite{yu2018generative} & 52.97\%\\
DLP-CNN~\cite{li2017reliable} & 54.47\%  \\
PG-CNN~\cite{li2018patch} & 55.33\%  \\
ResNet-PL~\cite{pan2019occluded} & 56.42\%  \\
gACNN~\cite{li2018occlusion} & 58.78\%  \\
\textbf{OADN(ours)}  & \textbf{61.89\%}  \\
\hline
\end{tabular}
\end{center}
\end{table}

\begin{table}
\caption{Validation set accuracy on Occlusion-AffectNet and Pose-AffectNet dataset}
\label{tab:affect_acc}
\begin{center}
\begin{tabular}{c|c|c|c}
\hline
Method & Occ. Acc. &  Pose\textgreater 30 Acc. & Pose\textgreater 45 Acc. \\
\hline
\hline
RAN~\cite{wang2020region} & 58.50\% & 53.90\% & 53.19\%\\
\hline
\textbf{OADN(ours)}  & \textbf{64.02}\% & \textbf{61.12}\% & \textbf{61.08}\%\\
\hline
\end{tabular}
\end{center}
\end{table}


We show the confusion matrix in Figure~\ref{fig:raf_cm}.
It is observed that \emph{Fear} and \emph{Disgust} are the two most confusing expression, where \emph{Fear} is easily confused with \emph{Surprise} because of similar facial appearance while \emph{Disgust} is mainly confused by \emph{Neutral} due to the subtleness of the expression.

\begin{figure}[!ht]
\centering
\includegraphics[width=0.45\textwidth]{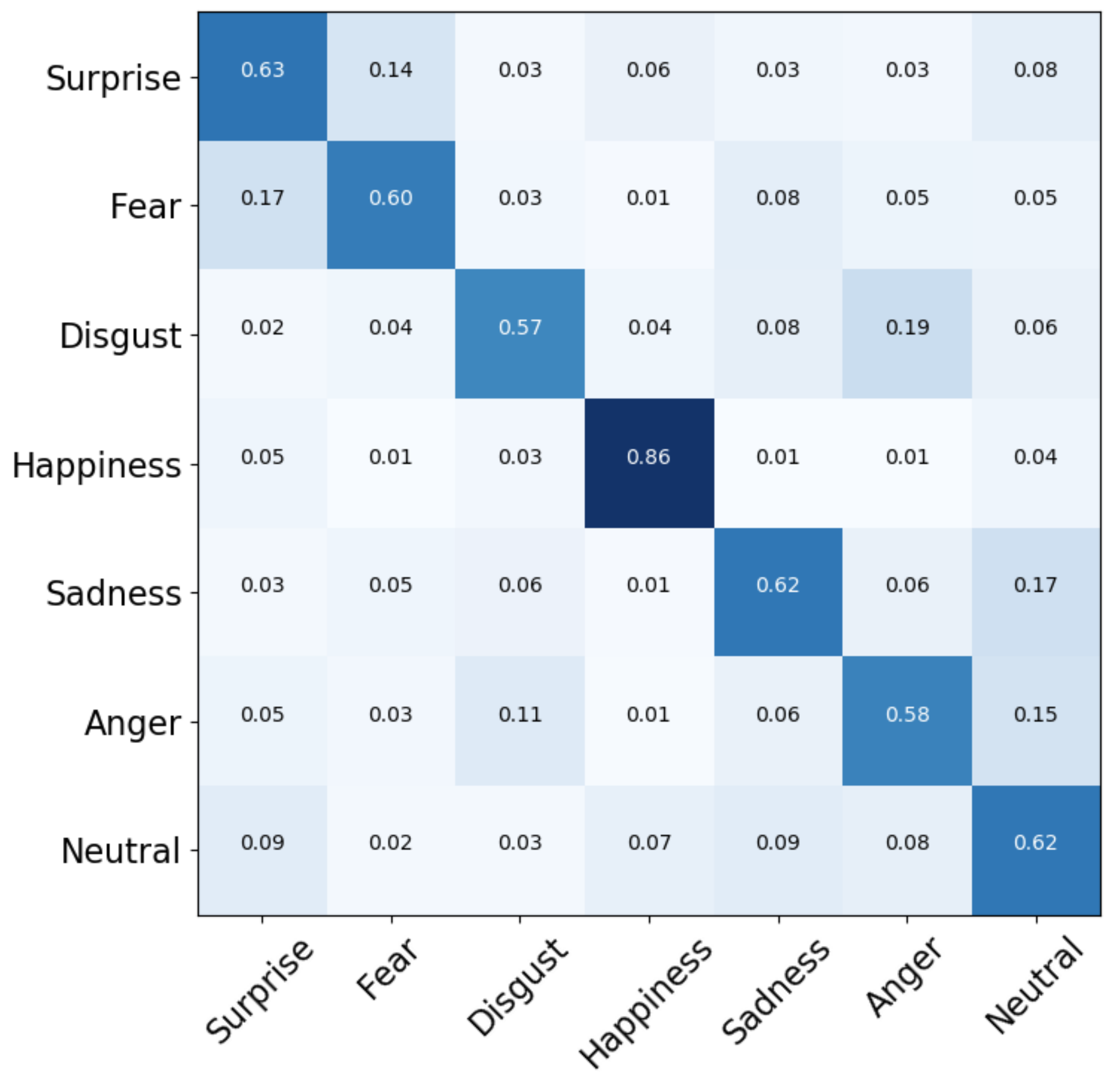}
\caption{Confusion matrix for Affectnet dataset. The darker the color, the higher the accuracy.}
\label{fig:affectnet_cm}
\end{figure}

\textbf{AffectNet}~\cite{mollahosseini2017affectnet} is currently the largest expression dataset. There are about 400,000 images manually annotated with seven discrete facial expressions and the intensity of valence and arousal. Following the experiment setting in~\cite{li2018occlusion}, we only used images with neutral and six basic emotions, containing 280,000 images for training and 3,500 images from the validation set for testing since the test set is not publicly available. Very recently, Wang \etal~\cite{wang2020region} released the \textbf{Occlusion-AffectNet} and \textbf{Pose-AffectNet} datasets where only images with challenging conditions are selected as the test sets. For the Occlusion-Affectnet, each image is occluded with at least one type of occlusion: wearing mask, wearing glasses, \emph{etc}.
There are a total of 682 images. For the Pose-AffectNet, images with pose degrees larger than 30 and 45 are collected. The number of images are 1,949 and 985, respectively.

As shown in Table~\ref{tab:affect_acc}, OADN achieves the best performance with an accuracy of 61.89\% on the validation set. Compared to the strongest competing method in the same setting gACNN~\cite{li2018occlusion}, OADN surpasses it by 3.1\%, which is a reasonable improvement. OADN also outperforms RAN~\cite{wang2020region} by 4.56\%, when both are pre-trained on a large-scale face recognition dataset.
On the Occlusion-AffectNet and Pose-AffectNet datasets, the performance gap between OADN and RAN is further increased. As a comparison, OADN exceeds RAN by 5.52\%, 7.22\% and 7.89\% on the test sets with occlusion, pose degree greater than 30 and 45, respectively.
This validates the effectiveness of the proposed method for the occluded facial expression recognition problem.
The confusion matrix is shown in Figure~\ref{fig:affectnet_cm}. From it we can find both \emph{Disgust} and \emph{Anger} are the most difficult expressions to classify.


\begin{table}
\caption{Test set accuracy on FED-RO dataset}
\label{tab:fed_acc}
\begin{center}
\begin{tabular}{c|c}
\hline
Method & Average Accuracy  \\
\hline
\hline
RAN~\cite{wang2020region} & 67.98\% \\
\textbf{OADN(ours)}  & \textbf{71.17\%} \\
\hline
VGG16~\cite{simonyan2014very} & 51.11\% \\
ResNet18~\cite{he2016deep} &64.25\%\\
GAN-Inpainting~\cite{yu2018generative} & 58.33\%\\
DLP-CNN~\cite{li2017reliable} & 60.31\%  \\
PG-CNN~\cite{li2018patch} & 64.25\%  \\
gACNN~\cite{li2018occlusion} & 66.50\%   \\
\textbf{OADN(ours)}  & \textbf{68.11\%}  \\
\hline
\end{tabular}
\end{center}
\end{table}

\textbf{FED-RO}~\cite{li2018occlusion} is a recently released facial expression dataset with real world occlusions. Each image has natural occlusions including sunglasses, medical mask, hands and hair. It contains 400 images labeled with seven expressions for testing. We train our model on the joint training data of RAF and AffectNet, following the protocol suggested in~\cite{li2018occlusion}.

\begin{figure}[!ht]
  \centering
  \includegraphics[width=0.45\textwidth]{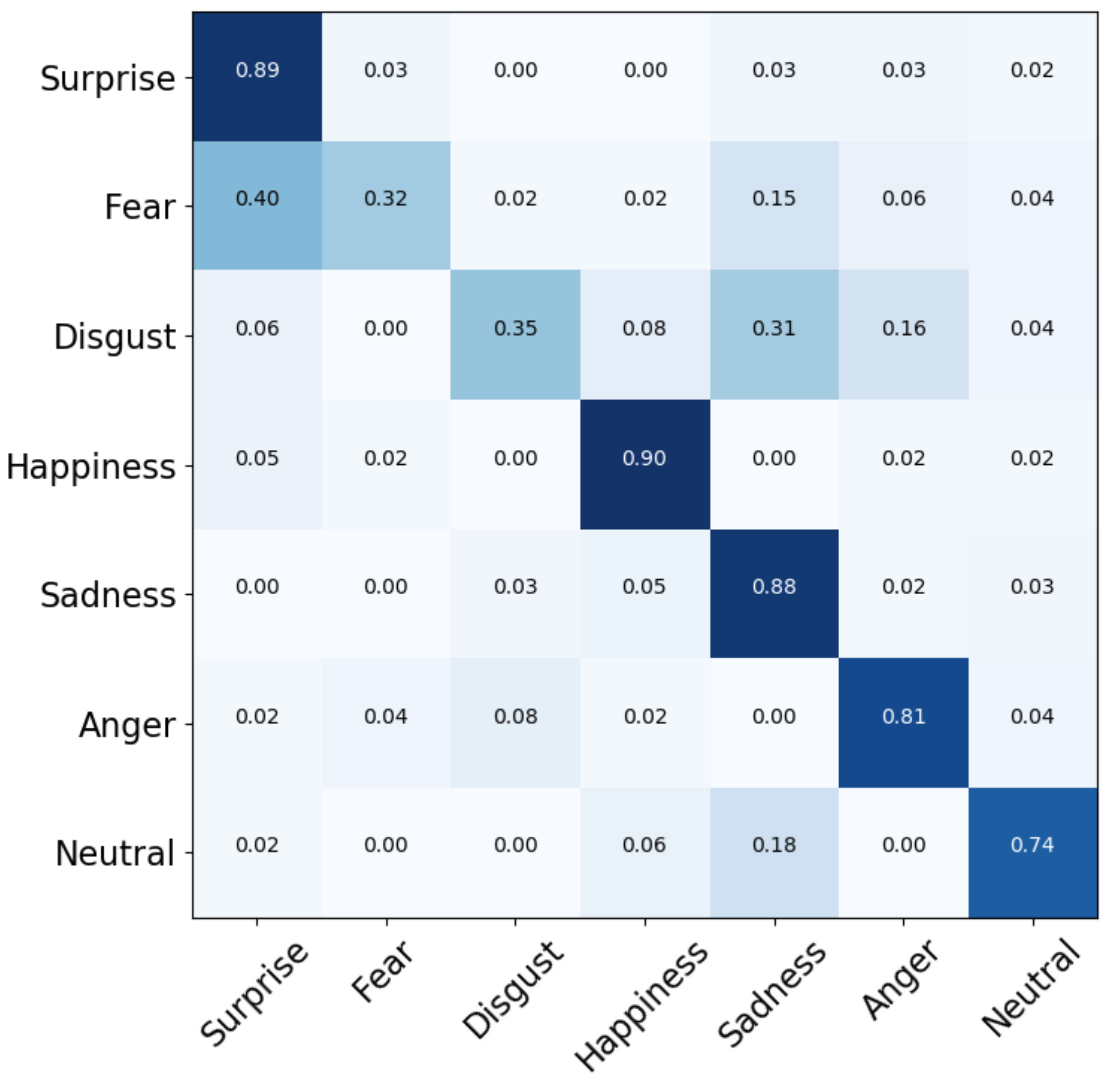}
  \caption{Confusion matrix for FED-RO dataset. The darker the color, the higher the accuracy.}
  \label{fig:fed_cm}
\end{figure}

As shown in Table~\ref{tab:fed_acc}, OADN achieves the best performance with an accuracy of 68.11\%, improving over gACNN by 1.61\%. 
OADN also outperforms RAN by 3.19\% when pre-trained on a face recognition dataset.
From the confusion matrix shown in Figure~\ref{fig:fed_cm}, we can see both \emph{Surprise} and \emph{Happy} have high accuracy, while \emph{Fear} and \emph{Disgust} are easily confused with \emph{Surprise} and \emph{Sad}.

\textbf{FERPlus}~\cite{barsoum2016training} is a real-world facial expression dataset initially introduced in ICML 2013 Challenge~\cite{goodfellow2013challenges}. It consists of 28,709 training images, 3,589 validation images and 3,589 test images. Each image is labeled with one of the eight expressions by 10 independent taggers. Recently, Wang \etal~\cite{wang2020region} released the \textbf{Occlusion-FERPlus} and \textbf{Pose-FERPlus} datasets, where images under occlusion and large pose (\textgreater30 and \textgreater45) are collected from the FERPlus test sets. 
The Occlusion-FERPlus has a total number of 605 images, while Pose-FERPlus has 1,171 and 634 images with pose larger than 30 and 45 degrees, respectively. 
Following~\cite{wang2020region}, we trained our model on the training data of FERPlus and test on these challenging datasets.

Table~\ref{tab:fer_acc} reports the test accuracy. The OADN significantly surpasses RAN by a large margin with 6.29\% and 7.10\% improvements on the Pose-FERPlus datasets. OADN also achieves better performance on the Occlusion-FERPlus dataset. This validates the effectiveness of our method for recognizing facial expressions under challenging conditions.

\begin{table}
\caption{Test set accuracy on Occlusion-FERPlus and Pose-FERPlus dataset}
\label{tab:fer_acc}
\begin{center}
\begin{tabular}{c|c|c|c}
\hline
Method & Occ. Acc. &  Pose\textgreater30 Acc. & Pose\textgreater45 Acc. \\
\hline
\hline
RAN~\cite{wang2020region} & 83.63\% & 82.23\% & 80.40\%\\
\hline
\textbf{OADN(ours)}  & \textbf{84.57}\% & \textbf{88.52}\% & \textbf{87.50}\%\\
\hline
\end{tabular}
\end{center}
\end{table}

\subsection{Ablation Study}
In this section, we conduct ablation studies on the RAF dataset to analyze each component of OADN.

\begin{figure}[!ht]
  \centering
  \includegraphics[width=0.5\textwidth]{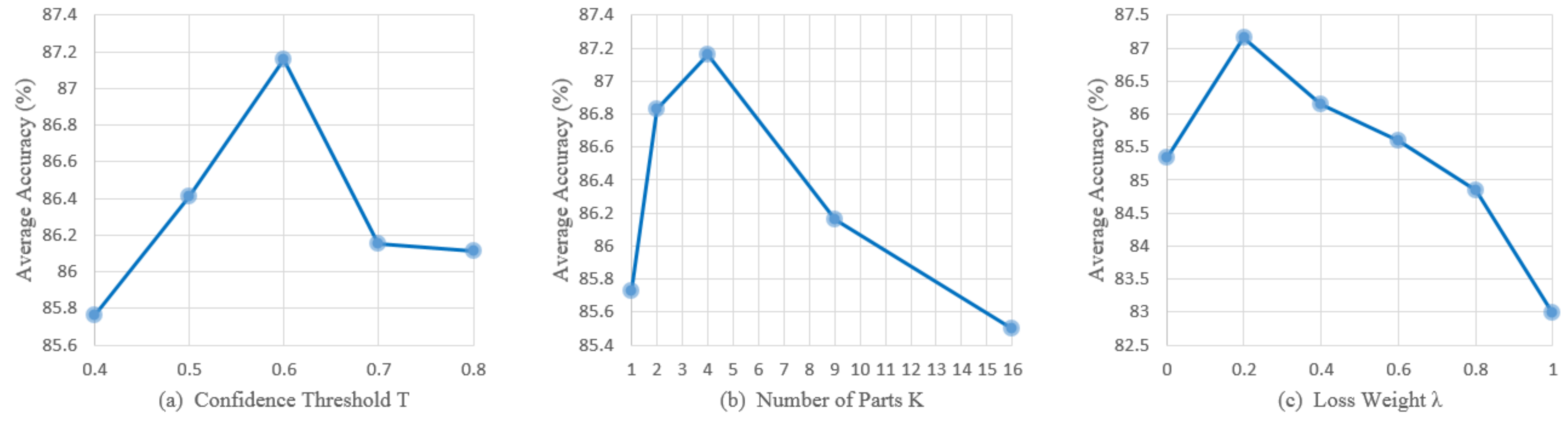}
  \caption{The impacts of the confidence threshold $T$, number of regions $K$ and the loss combination weight $\lambda$ on the performance of OADN.}
  \label{fig:ablation}
\end{figure}

\textbf{The impact of the landmark confidence threshold $T$.}
The confidence scores of the interest points are utilized to select the interest points from non-occluded facial areas. From Equation~(1), points with confidence scores higher than $T$ are kept. We can see from Figure~\ref{fig:ablation} (a) that with $T=0.6$, OADN achieves the best performance. When $T$ is further increased, the performance drops quickly since some important facial areas which may not be occluded are also thrown away. On the other hand, when $T$ becomes less than 0.6, OADN starts to perform worse. This is because noisy information from the occluded areas are also included, which degrades the clean features. 


\textbf{The impact of the number of regions $K$.}
In the facial region branch, we partition the global feature maps into $K$ blocks and train an expression classifier from each block independently. So $K$ decides the granularity of the part-level features. From Figure~\ref{fig:ablation} (b), it is observed that the best accuracy is achieved at $K=4$. When $K=1$, the facial region branch equals to the standard ResNet50 classifier. The worse performance indicates the necessity to learn features at part-level. However, increasing $K$ to be a large number like 16 does not bring further increasement. This is because when the facial region is too small, it lacks enough information to make the prediction due to the occlusion. Thus the classifiers are confused and the training is stagnated.

\textbf{The impact of the loss combination weight $\lambda$}.
To train OADN, we jointly optimize the loss from the landmark-guided attention branch (LAB) and the facial region branch (FRB) as defined in Equation~(5). The loss weight $
\lambda$ controls the relative importance of each loss. When $\lambda$ equals 1, only LAB is utilized. While $\lambda=0$ means only FRB is used. From Figure~\ref{fig:ablation} (c), we can find that LAB obtains better performance since the network is guided to attend to the most discriminative facial areas. While combining the two branches achieves better performance than using either one branch alone. This validates the effectiveness of the complementary features learned by the two branches.



\subsection{Visualization}
\begin{figure}[!ht]
\centering
\includegraphics[width=0.46\textwidth]{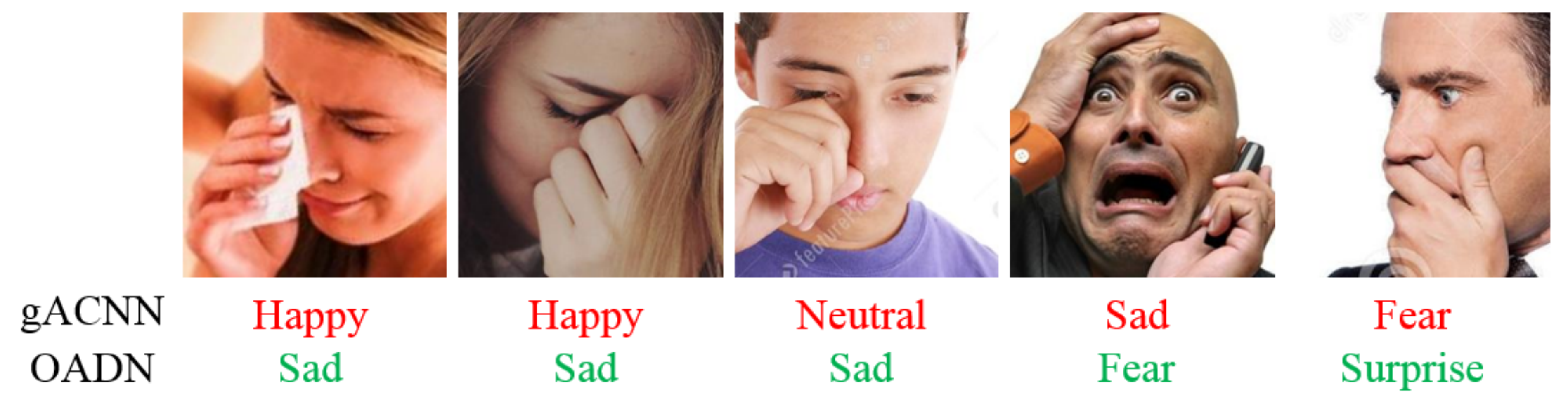}
\caption{Comparison of the gACNN method and our OADN method on the FED-RO dataset. Red and green texts indicate the error and correct predictions.}
\label{fig:vis}
\end{figure}

Figure~\ref{fig:vis} shows some expression recognition examples of the gACNN~\cite{li2018occlusion} and our OADN method on the FED-RO dataset. 
The classification results show that gACNN is vulnerable to large head poses and heavy facial occlusions. 
On the contrary, the OADN can work successfully under these conditions.

\section{Conclusions}
In this paper, we present an occlusion-adaptive deep network to tackle the occluded facial expression recognition problem. The network is composed of two branches: the landmark-guided attention branch guides the network to learn clean features from the non-occluded facial areas. The facial region branch increases the robustness by dividing the last convolutional layer into several part classifiers. 
We conduct extensive experiments on both challenging in-the-wild expression datasets and real-world occluded expression datasets.
The results show that our method outperforms existing methods and achieves robustness against occlusion and various poses.

\section{Acknowledgement}
This research is based upon work supported by the Office
of the Director of National Intelligence (ODNI), Intelligence
Advanced Research Projects Activity (IARPA), via IARPA
R\&D Contract No. 2019-022600002. The views and conclusions
contained herein are those of the authors and should not
be interpreted as necessarily representing the official policies
or endorsements, either expressed or implied, of the ODNI,
IARPA, or the U.S. Government. The U.S. Government is authorized
to reproduce and distribute reprints for Governmental
purposes notwithstanding any copyright annotation thereon.

{\small
\bibliographystyle{ieee}
\bibliography{egbib}
}

\end{document}